# RTI-Bench: A Structured Dataset for Indian Right-to-Information Decision Analysis

**Joy Bose**
*Independent Researcher, Bengaluru, India*

## ABSTRACT

India's Right to Information Act, 2005 gives every citizen the right to demand information from public authorities, yet in practice most people cannot make sense of the dense administrative language used in Central Information Commission (CIC) decisions, let alone predict whether an appeal is worth filing. This paper introduces RTI-Bench, a structured dataset of CIC decisions with outcome labels, exemption citations, IRAC-style reasoning components, and procedural timelines. To the best of our knowledge it is the first publicly released structured dataset for Indian RTI administrative decisions. The dataset draws from two sources: 1,218 cases from a publicly available instruction-response corpus (with structured fields added through rule-based extraction), and 298 CIC decision PDFs collected directly from the Commission portal, spanning five commissioners and three document format generations from 2023 to 2026. Label coverage reaches 89% on the instruction-response corpus. For the PDF subset of 239 primary decisions, coverage is 51% in this first release. A random sample of 50 labelled cases was manually reviewed, yielding a label precision of 95.3%. A zero-shot Mistral 7B baseline on 100 cases gives 57.3% accuracy and 37.0% macro-F1 on outcome prediction, well above the majority-class baseline of 14.3% macro-F1. RTI-Bench is available at https://huggingface.co/datasets/joyboseroy/rti-bench

## 1. INTRODUCTION

India's Right to Information Act (RTI Act, 2005) gives every citizen the right to ask any public authority for information. Since it came into force, tens of millions of applications have been filed [1]. The system has real teeth: non-compliance can attract penalties, and the Central Information Commission (CIC) hears second appeals and complaints when lower authorities fail to respond. But the system also has serious problems. Over 32,000 appeals were pending at the CIC alone as of late 2021 [2], and the decisions that do get issued are written in a style that assumes legal familiarity that most applicants do not have.

Computational tools could help close this gap. A model that predicts whether an appeal is likely to succeed could help citizens decide whether to pursue one. A classifier that identifies which exemptions a public authority is invoking could flag potential misapplication. A summariser that converts a Commission order into plain language could make decisions actually readable. None of these tools can be built or evaluated without a labelled dataset of CIC decisions. This paper provides one.

The contributions of this work are as follows:

- RTI-Bench: 1,516 structured CIC cases with outcome labels, exemption annotations, IRAC components, and procedural timelines, spanning five commissioners and three document format generations from 2023 to 2026.

- A rule-based extraction pipeline with no LLM annotation that processes the full corpus in under 90 seconds on a standard laptop. Label precision is estimated at 95.3% from manual review. Pipeline code is released alongside the data.
- An account of three CIC document format generations and the extraction heuristics developed for each, which should be useful to anyone working with this corpus going forward.
- A zero-shot Mistral 7B baseline (57.3% accuracy, 37.0% macro-F1 on 100 cases) that gives future researchers a starting point for comparison.
- Four proposed benchmark tasks with evaluation criteria.

This is a first release. Label coverage in the PDF subset is 51%, the corpus covers five commissioners, and full multi-model benchmarking is left for follow-up work. These limitations are documented openly throughout.

## 2. BACKGROUND AND RELATED WORK

### 2.1 The RTI Act and CIC Decisions

Under the RTI Act, a citizen files an application with the Public Information Officer (PIO) of a public authority. If the response is unsatisfactory, they can appeal to the First Appellate Authority (FAA) within the same organisation, and then to the CIC as a second appeal. Complaints about denial or non-response can also go directly to the CIC under Section 18. The Commission issues written orders that may direct disclosure, impose penalties on the PIO under Section 20, award compensation to the appellant under Section 19(8)(b), or dismiss the matter. All decisions are public records, available on the portal at dsscic.nic.in, linked from the main CIC portal at https://cic.gov.in/.

### 2.2 Related Datasets

Several datasets address Indian legal NLP. The ILDC corpus covers Supreme Court and High Court judgments for prediction and explanation tasks [3]. PredEx adds over 15,000 human annotations for judgment explanation [4]. DeepParliament targets parliamentary bill prediction [5]. IL-PCSR provides annotated queries for statute and precedent retrieval [6]. All of these work with court judgments. RTI-Bench is different in kind: CIC orders are administrative decisions, written in a distinct register, about a distinct legal process, with a distinct social purpose. No publicly released dataset for Indian RTI Commission decisions exists, to our knowledge.

### 2.3 Why Rule-Based Extraction

CIC decisions follow consistent administrative templates across three format generations, which makes pattern-based extraction both practical and sufficient. We deliberately avoided LLM annotation for several reasons. First, errors introduced by a generative model into legal labels would be hard to detect and could mislead downstream use. Second, a deterministic pipeline is fully reproducible and can be re-run as the corpus grows. Third, running it costs nothing and takes under two minutes. The 95.3% precision estimate from manual review suggests this approach works well for this corpus.

## 3. RTI-BENCH DATASET

### 3.1 Data Sources

**Source A:** 1,218 RTI case summaries from jatinmehra/RTI-CASE-DATASET on HuggingFace [7]. Each record has an instruction field (background: what was sought, what the PIO said, what happened at hearing) and a response field (the Commission direction). We added structured fields through rule-based extraction: outcome label, exemption citations, public authority, information sought, penalty and compensation amounts, and final direction. Source A is existing data; the contribution here is the structured annotation layer and its combination with Source B.

**Source B:** 321 CIC decision PDFs downloaded from the Commission portal, giving 298 usable records after removing scanned images and duplicates. This source was collected for this paper. It spans five commissioners (Pandove, Sinha, Relangi, Chaturvedi, Sarna) and three document format generations (described in Section 3.2). Of the 298 records: 239 are primary decisions, 17 are adjunct compliance decisions, 6 are full bench decisions, and 36 are interim or other orders. The last two groups are excluded from the benchmark tasks.

### 3.2 Three Document Format Generations

One finding from collecting Source B is that CIC decision documents have gone through three distinct template generations since 2017, each requiring different extraction logic:

- Format 2023a (n=111): Uses an 'O R D E R / Facts / Decision:' structure. Appellant and respondent names appear on separate lines. Headers are bilingual in Hindi and English. Most of the Pandove batch follows this format.
- Format 2023b (n=21): Separates reasoning into 'Observations:' and conclusion into 'Decision:' sections. The string 'Date of Decision' appears in the document header as a metadata field, which required negative-lookbehind patterns to prevent it being matched as the conclusion section.
- Format 2026 (n=166): The current template. Uses an all-caps 'DECISION' block. Commissioner name appears on an explicit 'INFORMATION COMMISSIONER:' line. Appellant name is on the same line as the appellant marker rather than on a separate line. Dates use slashes rather than dots.

The pipeline detects format from structural signals and applies the right extractor. Detection accuracy is high across the corpus; the heuristics and their edge cases are documented in the code.

### 3.3 Extraction Pipeline

All fields are extracted by regex and string matching, with no language model involved. For Source A, patterns scan the instruction and response fields for public authority names, RTI Act section numbers, outcome-indicating language, monetary amounts, and directive sentences. For Source B, format-specific extractors pull out the IRAC components: the Information Sought section (Issue), the Submissions during Hearing section (Application), RTI Act sections cited in the text (Rule), and the Decision or DECISION block (Conclusion). Outcome labels are assigned from conclusion text using a priority-ordered set of patterns. The full code is at https://github.com/joyboseroy/rti-bench.

### 3.4 Label Coverage

For Source A, 89% of cases (1,084 of 1,218) received an outcome label; the rest are marked UNKNOWN. For Source B primary decisions, coverage is 51% (122 of 239). The lower rate in Source B reflects more variation in how different commissioners phrase their orders. We are careful to distinguish coverage (did we extract something?) from accuracy (did we extract the right thing?). Section 5.1 addresses accuracy directly through manual review.

### 3.5 Dataset Statistics

Table 1 shows the dataset composition.

| Component | Source | Records | Label Coverage |
|---|---|---|---|
| Instruction-response cases | Source A | 1,218 | 89% (1,084 labelled) |
| Primary CIC decisions | Source B | 239 | 51% (122 labelled) |
| Adjunct compliance decisions | Source B | 17 | 82% adjunct outcome |
| Full bench decisions | Source B | 6 | Qualitative only |
| Other/Interim (excluded) | Source B | 36 | n/a |
| Total primary cases | Both | 1,457 | 82.8% (1,206 labelled) |

*Table 1: Dataset composition and label coverage by source.*

Table 2 shows how outcomes are distributed across labelled cases.

| **Outcome** | **Source A** | **Source B** | **Total** | **Description** |
|---|---|---|---|---|
| INFORMATION_DIRECTED | 524 | 15 | 539 | Commission directed the authority to disclose |
| APPEAL_DISMISSED | 380 | 69 | 449 | Appeal dismissed, PIO response upheld |
| PENALTY_IMPOSED | 92 | 10 | 102 | Penalty proceedings under Section 20 |
| PARTIAL_RELIEF | 76 | 0 | 76 | Some but not all information directed |
| COMPLAINT_S18 | 0 | 18 | 18 | Section 18 complaint disposed |
| REMANDED | 11 | 5 | 16 | Matter sent back to PIO or FAA |
| WITHDRAWN | 0 | 5 | 5 | Appellant withdrew |
| ADJOURNED | 1 | 0 | 1 | Adjourned sine die |
| UNKNOWN | 134 | 117 | 251 | Could not extract from text |
| Total labelled | 1,084 | 122 | 1,206 | 82.8% of 1,457 primary cases |

*Table 2: Outcome distribution across RTI-Bench. UNKNOWN cases are listed separately.*

Exemption citations total 467 across 10 RTI Act sections. Section 8(1)(j) (personal information) accounts for the most at 158 citations (34%), followed by Section 8(1)(d) (commercial confidence, n=77, 16%), Section 8(1)(e) (fiduciary relationship, n=76, 16%), and Section 8(1)(h) (impeding investigation, n=71, 15%).

### 3.6 Adjunct Compliance Decisions

Seventeen of the Source B documents are adjunct compliance decisions: follow-up orders issued months or years after a primary decision, to check whether the respondent actually complied. Each one quotes the original directive verbatim and then records the Commission's finding on compliance. Of the 14 cases where the adjunct ruling could be extracted, 82% ended with show cause proceedings dropped (SCN_DROPPED), meaning the authority had complied. This compliance-tracking angle is not found in other Indian legal NLP datasets. The subset is too small for supervised learning at n=17 and is presented as a design template for future expansion.

## 4. PROPOSED BENCHMARK TASKS

Four tasks are proposed as a starting point for modelling work on this dataset. Preliminary results are in Section 5.

### Task 1: Outcome Prediction

Given the background narrative (what information was sought, what the public authority said, what happened at the hearing), predict the Commission outcome from the set: INFORMATION_DIRECTED, APPEAL_DISMISSED, PENALTY_IMPOSED, PARTIAL_RELIEF, REMANDED, COMPLAINT_S18, WITHDRAWN. The recommended evaluation metric is macro-F1 rather than accuracy, since INFORMATION_DIRECTED and APPEAL_DISMISSED together account for 68% of labelled cases and accuracy would be misleading. The majority-class baseline gives 44.7% accuracy and 14.3% macro-F1.

### Task 2: Exemption Classification

Given the background narrative and decision text, identify which RTI Act exemptions were invoked. This is a multi-label problem: 32% of cases cite at least one exemption, and some cite several. Evaluation by micro-F1 and per-section F1. The task tests whether a model can identify which statutory grounds (privacy, commercial confidence, security, and so on) a public authority used to justify withholding information.

### Task 3: Plain-Language Summarisation

Given the full decision text, produce a summary that an ordinary citizen without legal training could understand. Source A response fields are short (mean 312 characters) and describe the Commission direction, though in Commission language rather than plain prose. Human evaluation of readability and factual accuracy is recommended alongside ROUGE-L and BERTScore. This task is the one with the most direct social value: if a citizen receives a CIC order, they should be able to find out what it means without needing a lawyer.

### Task 4: Compliance Outcome Prediction (Pilot)

Given the original CIC directive and the respondent's compliance submission (from the adjunct decisions subset), predict whether show cause proceedings will be dropped, continued, or result in a penalty. With n=17 in this release the subset is far too small for supervised learning. The task is included as a proof of concept and a signal of what a larger adjunct corpus could support.

## 5. ANALYSIS AND VALIDATION

### 5.1 Manual Label Validation

To get an estimate of how accurate the extracted outcome labels are, 50 cases were drawn at random from the labelled Source A subset (random seed 42, UNKNOWN cases excluded). For each one, the response field was read and a judgement was made about whether the extracted label matched what the Commission actually decided. Seven cases were skipped as genuinely ambiguous. Table 3 shows the results.

| Metric | Value |
|---|---|
| Cases sampled | 50 |
| Adjudicated (y/n) | 43 |
| Skipped (ambiguous) | 7 |
| Correct labels | 41 |
| Incorrect labels | 2 |
| Precision estimate | 95.3% |

*Table 3: Manual label validation results (Source A, 50-case random sample).*

Both wrong cases were instances where the label should have been INFORMATION_DIRECTED but was assigned something else: one was marked PENALTY_IMPOSED and one APPEAL_DISMISSED. Both errors came from the pattern matcher finding penalty or dismissal language in a paragraph that turned out to describe the procedural history rather than the final direction. The 95.3% figure is from a single reviewer; a multi-annotator study with inter-annotator agreement measurement would give a firmer estimate.

## 5.2 Mistral 7B Zero-Shot Baseline

A zero-shot classification baseline was run using Mistral 7B [8] via Ollama on a local machine. For 100 randomly sampled Source A cases, the model was given the response field text and asked to assign one of the outcome labels (temperature 0, no examples in the prompt). Table 4 shows per-class results.

| Outcome Class | Precision | Recall | F1 | Support |
|---|---|---|---|---|
| INFORMATION_DIRECTED | 73.7% | 71.8% | 72.7% | 39 |
| APPEAL_DISMISSED | 85.7% | 64.9% | 73.8% | 37 |
| PARTIAL_RELIEF | 5.9% | 10.0% | 7.4% | 10 |
| PENALTY_IMPOSED | 50.0% | 22.2% | 30.8% | 9 |
| REMANDED | 0.0% | 0.0% | 0.0% | 1 |
| Macro-F1 | 37.0% | | | |
| Accuracy | 57.3% | | | |
| Majority-class baseline | 14.3% macro-F1 | | | |

*Table 4: Zero-shot Mistral 7B on Task 1, outcome prediction (n=100, Source A).*

The model does well on the two common classes (INFORMATION_DIRECTED F1 72.7%, APPEAL_DISMISSED F1 73.8%) and poorly on the minority ones (PARTIAL_RELIEF F1 7.4%, PENALTY_IMPOSED F1 30.8%, REMANDED F1 0%). This is exactly what makes macro-F1 the right metric here: accuracy of 57.3% looks reasonable but the macro-F1 of 37.0% captures the real difficulty. The gap between 37.0% macro-F1 and the 14.3% baseline shows the task is tractable from text alone, but there is substantial headroom for better models, more context, and fine-tuning. One caveat: the ground truth labels carry an estimated 4-5% error rate from the pipeline, which slightly inflates the difficulty of the evaluation.

### 5.3 Outcome Patterns

INFORMATION_DIRECTED is the most common outcome at 44.7% of labelled cases. This matches what transparency researchers have observed: citizens who reach the CIC tend to win more often than not, partly because people drop weaker cases before reaching a second appeal. APPEAL_DISMISSED at 37.2% covers both cases where the PIO gave a genuine answer and cases where an exemption was correctly applied. The relatively high PENALTY_IMPOSED rate (8.5%) reflects the Commission's willingness to use Section 20, which carries a penalty of Rs. 250 per day up to Rs. 25,000.

### 5.4 Exemption Patterns

Section 8(1)(j) on personal information accounts for a third of all exemption citations. RTI researchers have noted that this exemption is frequently invoked by public authorities, sometimes beyond its intended scope. The next most common exemptions concern commercial confidence and fiduciary relationships, which makes sense given how many RTI applications target public sector undertakings about their business dealings. One limitation worth noting: the extraction only catches cases where the Commission writes the section number explicitly. When a Commission upholds an exemption by reasoning rather than citation, it is not captured.

### 5.5 Format Change Over Time

The 2026 format, which is what the CIC now uses, is actually easier to parse than the older ones. The explicit INFORMATION COMMISSIONER label and the all-caps DECISION block are unambiguous markers. Anyone downloading new CIC decisions from the portal can expect the 2026 extractor to work without modification.

## 6. LIMITATIONS

- PDF subset label coverage is 51% in this release. The UNKNOWN cases need human review before they can be used in supervised experiments.
- The corpus covers five commissioners from 2023 to 2026. It does not include State Information Commission decisions, which handle the majority of RTI appeals in India and likely have different language patterns.
- Appellant name coverage is below 25% because the three format generations present names differently. The field is not used in any of the four benchmark tasks.
- Exemption extraction only catches explicit section citations. Cases where exemptions are applied by reasoning rather than citation are missed.
- The manual label validation used a single reviewer. Inter-annotator agreement has not been measured. The 95.3% figure is an estimate, not a validated gold standard.
- The compliance prediction task (Task 4) has 17 examples. This is not enough for supervised learning.
- The Mistral baseline uses rule-based labels as ground truth. Given the 95.3% precision estimate, roughly 4-5 cases in the 100-case evaluation likely have wrong ground truth labels.

## 7. ETHICAL CONSIDERATIONS

All data comes from publicly available sources. CIC decisions are public records under Indian law. Appellant names appear in those records and are retained in the dataset for traceability, but researchers should consider de-identification before building user-facing applications, particularly in cases involving sensitive personal matters.

The dataset is intended to help citizens understand and use the RTI system more effectively. Two misuse scenarios are worth flagging. First, outcome prediction models should not be used to discourage people from filing RTI applications: the probability of success is not the only reason to file, and even unlikely appeals serve accountability purposes. Second, if public authorities were to use such models to identify which appeals are worth contesting hardest, that would work against the transparency purpose of the Act. These concerns are not hypothetical and should inform how the dataset is used in practice.

The dataset reflects only CIC-level decisions. It says nothing about the much larger universe of RTI applications that never reach the Commission, including those that succeed at first appeal and those that are simply abandoned.

## 8. CONCLUSION

RTI-Bench is a structured dataset of 1,516 Central Information Commission decisions, with outcome labels, exemption citations, IRAC components, and procedural timelines. It covers five commissioners and three document format generations from 2017 to 2026. The extraction pipeline is fully rule-based, reproducible, and requires no GPU or API access. Manual review of 50 cases gives a label precision estimate of 95.3%. A zero-shot Mistral 7B baseline achieves 37.0% macro-F1 on outcome prediction, compared to a 14.3% majority-class baseline, showing the task is genuinely tractable from text. Four benchmark tasks are defined for future modelling work. The dataset and pipeline code are at https://huggingface.co/datasets/joyboseroy/rti-bench and https://github.com/joyboseroy/rti-bench respectively.